\documentclass{article}

\usepackage{PRIMEarxiv}

\usepackage[utf8]{inputenc} % allow utf-8 input
\usepackage[T1]{fontenc}    % use 8-bit T1 fonts
\usepackage{amsmath}
\usepackage{hyperref}       % hyperlinks
\usepackage{url}            % simple URL typesetting
\usepackage{booktabs}       % professional-quality tables
\usepackage{amsfonts}       % blackboard math symbols
\usepackage{nicefrac}       % compact symbols for 1/2, etc.
\usepackage{microtype}      % microtypography
\usepackage{lipsum}
\usepackage{fancyhdr}       % header
\usepackage{graphicx}       % graphics
\graphicspath{{media/}}     % organize your images and other figures under media/ folder

\title{Understanding writing style in social media with a supervised contrastively pre-trained transformer
}

\author{
  Javier Huertas-Tato, Alejandro Martín, David Camacho \\
  Departamento de Sistemas Informáticos \\
  Universidad Politécnica de Madrid \\
  Madrid\\
  \texttt{\{javier.huertas.tato, alejandro.martin, david.camacho\}@upm.es} \\
}

\begin{document}
\maketitle

\begin{abstract}
Online Social Networks serve as fertile ground for harmful behavior, ranging from hate speech to the dissemination of disinformation. Malicious actors now have unprecedented freedom to misbehave, leading to severe societal unrest and dire consequences, as exemplified by events such as the Capitol assault during the US presidential election and the Antivaxx movement during the COVID-19 pandemic. Understanding online language has become more pressing than ever. While existing works predominantly focus on content analysis, we aim to shift the focus towards understanding harmful behaviors by relating content to their respective authors. Numerous novel approaches attempt to learn the stylistic features of authors in texts, but many of these approaches are constrained by small datasets or sub-optimal training losses. To overcome these limitations, we introduce the Style Transformer for Authorship Representations (STAR), trained on a large corpus derived from public sources of $4.5 \cdot 10^6$ authored texts involving 70k heterogeneous authors. Our model leverages Supervised Contrastive Loss to teach the model to minimize the distance between texts authored by the same individual. This author pretext pre-training task yields competitive performance at zero-shot with PAN challenges on attribution and clustering. Additionally, we attain promising results on PAN verification challenges using a single dense layer, with our model serving as an embedding encoder. Finally, we present results from our test partition on Reddit. Using a support base of 8 documents of 512 tokens, we can discern authors from sets of up to 1616 authors with at least 80\% accuracy. We share our pre-trained model at huggingface \href{https://huggingface.co/AIDA-UPM/star}{AIDA-UPM/star} and our code is available at \href{https://github.com/jahuerta92/star}{jahuerta92/star}
%% Link to github
%% Link to transformers
\end{abstract}

% keywords can be removed
\keywords{Transformers \and Online Social Networks \and Style embeddings \and}

\section{Introduction} %% Main contribution figure
%\textbf{ESQUEMA}
Disinformation, hate speech, sexism, vital social issues spread at an unprecedented rate in Online Social Networks (OSN). Harmful content is a large-scale safety hazard as demonstrated by the COVID-19 antivaxx movement, the USA capitol riots among other events.
Users can target public figures for harassment, unwittingly collaborate in spreading misinformation or coordinate hateful groups with the help of OSN. They use public spaces for their own ends, polluting OSNs with large amounts of harmful content. These dangerous messages are left online, for the general public to see and interact. 

Frequent approaches for combating harmful content has been to directly detect whether a piece of information is true or false, harmful or harmless, offensive or funny.
This allows for content moderators to quickly label and remove all content that may be problematic. This approach has been studied extensively,
allowing for better content moderation, however this approach has disadvantages. First, the authors that publish content may repeat their behaviour several times before they are identified, leaving amounts of harmful content published. Second, it requires a reaction, harmful behaviours always have an inherent advantage in this regard. Finally, restricting exclusively content, while quick, requires individual consideration of each piece. Our proposal is to swerve research towards the authors of harmful content, aiming to profile their stylistic nuance for early detection for harmful content. While evading restrictions is a common practice in Online Social Networks (OSNs), it's important to note that the same author is often behind these actions. Therefore, identifying their unique rhetorical style becomes crucial. Thus, our aim is to characterize an author using exclusively the textual content they produce. 

If text is meant to be understood then transformers lead the state of the art. Popular Natural Language Processing tasks harness transformers to achieve incredible results, showing high capabilities on a wide arrangement of tasks.
Observing the success of these models at understanding content and semantics we propose that they can also achieve high performance at understanding style and structure. When style and structure are to be analyzed, challenges usually are led by feature-extraction methods even in recent works~\cite{kestemont2019overview}.
This just means that transformers are not pre-trained with an objective for detecting these features, they look mostly at semantics and content. Our proposal is to use a transformer and pre-train it to characterize the style of an author, the same way transformers generate similarity embeddings after being tuned to paraphrase.

We hypothesize that, when an author writes a text they do it with a measurable style and structure, which can be analysed and compared against other pieces of written content from the same or other authors. Using techniques such as Contrastive Learning we can compare texts against positive examples (other sources from the same authors) and negative examples (sources from anyone else). This pretraining method will develop a hyper-space where texts can be projected and where texts written by the same author have similar representations.

\begin{figure}[h] 
\centering
\includegraphics[width=0.6\textwidth]{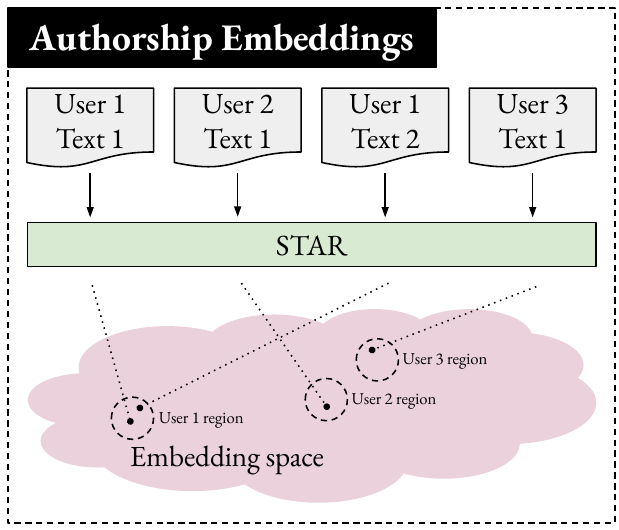}
\caption{STAR representations in context.}\label{fig:similarity}
\end{figure}

Hereby we present a Style Transformer for Authorship Representations (STAR), a pre-trained pure transformer approach to style and authorship understanding. Our contributions include the pretrained model and the pre-training method to develop our results, and we validate this model on two fronts: PAN-CLEF
challenges and over Reddit users~\cite{volske-etal-2017-tl}. 
A summary of how the model is meant to work is represented in Figure~\ref{fig:similarity} where 3 authors are evaluated.

Our document presents the current works in this field (Section ~\ref{sec:related}) to frame our methods (Section~\ref{sec:methods}), the model is tested using our validation methodology (Section~\ref{sec:setup}) and the results are presented and discussed (Section~\ref{sec:results}) finally we draw the conclusions in the last section along weaknesses and future work (Section~\ref{sec:conclusions})

\section{Related Work}\label{sec:related}
The proliferation of online media content represents a significant paradigm shift across various domains of knowledge. Within the realm of Natural Language Processing (NLP), there is an ongoing explosion of billions of textual pieces, generated at an accelerating pace. This abundance of authored texts empowers us to efficiently employ cutting-edge transformer models in the field of authorship analysis, leveraging resources such as blogs and digital libraries. In this section, we focus on delineating authorship within the context of our research and delve into methodologies that enable us to quantitatively capture the intricacies of style and authorship.

The act of writing is inherently influenced by the individual performing it. Their beliefs, knowledge, mannerisms, and other factors impact the content of any sufficiently extensive document. With this assumption in mind, we define authorship as a diverse set of identifiable textual features that are unique to a specific author. Authorship attribution, for example, seeks to establish connections between these stylistic features and the author~\cite{Stamatatos2009Mar}. Numerous fields of study, including linguistics, forensics, and history, approach authorship determination from various perspectives~\cite{juola2008authorship}. These fields stand to benefit from the development of accurate automated authorship models.

When considering attribution, two primary features of text come into play: content and style~\cite{sundararajan2018represents}. Stylometric features, such as word choice, frequency, punctuation, and sentence length, can be easily identified by computers~\cite{sari2018topic}. These features can be assembled into sets of representative characteristics, which can then be integrated into machine learning approaches like ensembles~\cite{custodio2018each}. The goal is to uncover differentiating attributes that contribute to authorship identification.

Authorship attribution applications span various domains, such as literature. Recent research applies a LDA-Transformer neural network to identify authorship in Chinese poetry~\cite{ai2021lda} and explores authorship in Russian literature, comparing classical machine learning models, deep learning recurrent neural networks, and transformers~\cite{fedotova2022authorship}. Complex techniques like Dynamic Authorship Attribution (DynAA) merge heterogeneous sources using a stack of classifiers for improved attribution~\cite{custodio2021stacked}. Generalization and domain bias challenges persist in authorship attribution, necessitating diverse datasets for evaluation~\cite{murauer2021developing, hitschler2017authorship, bevendorff2020overview}. Large texts introduce issues like co-authoring and style change detection~\cite{deibel2021style, Nieuwazny2021Sep}.

\subsection{Embeddings as style encoders}
Embeddings provide a useful and rich representation of stylistic, semantic and syntactic properties that can reveal authorship. Models such as PolitiBETO~\cite{villa2022nlp}, BETO~\cite{canete2020spanish}, and BERT~\cite{Devlin2018Oct}, may offer rich stylistic, semantic, and syntactic representations for authorship analysis. BERT-based models have been used in profiling irony and stereotype spreaders on Twitter~\cite{yu2022bert}, parallel stylometric document embeddings~\cite{math10050838}, and siamese networks~\cite{tyo2021siamese}. Authorship verification with BERT involves extracting embeddings from short text samples and unifying them using the median~\cite{futrzynski2021author}. Sentence-BERT is another transformer-based model employed for identifying hate speech spreaders~\cite{reimers2019sentence, schlicht2021unified}.

Researchers have evaluated embeddings generated by various methods. Kumar et al.~\cite{kumar2022comparing} compared Doc2Vec, GloVE, and FastText embeddings, finding FastText CBOW to perform best, albeit with higher resource usage. Another study~\cite{terreau2021writing} introduced an evaluation framework for author embeddings, focusing on writing style and noting the influence of semantic factors. In the state of the art, linguistic patterns are highlighted for authorship detection. A T5 model with CNN and attention mechanisms captures stylistic and semantic features~\cite{najafi2022text}. For author identification of anonymized papers~\cite{10.1145/3018661.3018735}, a meta path selection approach was employed in a bibliographic network. DeepStyle~\cite{10.1007/978-3-030-60290-1_17} focuses on stylistic information, using Triplet loss for social media post embeddings. Sentence syntactic structure is emphasized for authorship patterns~\cite{10.1145/3491203}, employing a semi-supervised model with lexical and syntactic sub-networks.

Transformers have quickly become essential in NLP, starting with Attention~\cite{Vaswani2017Jun} and BERT~\cite{Devlin2018Oct}. Numerous transformer models now exist for natural language comprehension, emphasizing semantic aspects with limited stylistic understanding. Works on style often prioritize content generation, e.g., Styleformer~\cite{Park2022}. Audio modalities have explored style~\cite{choi2020encoding}. Authorship interest has arisen within transformers for code generation~\cite{Kalgutkar2019Feb}.

We require encoder transformers like RoBERTa~\cite{liu2019roberta}, as decoder-style models yield less contextualized representations. For authorship, BertAA~\cite{fabien2020bertaa} is closest to our task, combining features and transformers. Embeddings, especially from RoBERTa, facilitate complex authorship tasks, even in challenging scenarios like social media microtext~\cite{suman2021authorship}. More advanced encoder models are viable but resource-intensive.

\subsection{Towards universal style embeddings}
We aim to combine transformers with contrastive approaches to tackle the authorship problem, developing a model to generate a representation of style encoded in a numerical embedding, as independent from the domain as possible. Recent advances demonstrate that self-supervised learning, seen in models like T5~\cite{Raffel2019Oct} and GPT-3~\cite{Brown2020May}, trained with methods like Causal Language Modeling (CLM), achieve state-of-the-art generalization, even in out-of-domain benchmarks. However, these methods are semantically-oriented. A promising alternative is Contrastive Learning, which allows direct representation learning without masking or pseudo-labels. In various domains, semi-supervised contrastive approaches have gained traction. For instance, CLIP~\cite{radford2021learning} aligns images and text for zero-shot classification, while SimCSE~\cite{gao2021simcse} employs contrastive learning for sentence similarity tasks. Authorship attribution relies on labels, and supervision can enhance contrastive learning~\cite{khosla2020supervised}, as seen in natural language inference~\cite{hu2022pair}. 

Finally, similar approaches based on contrastive learning have begun to emerge for style understanding, employing techniques such as triplet loss~\cite{hu2020deepstyle, wegmann-etal-2022-author} or InfoNCE~\cite{Huertas-Tato2022Jul}. Other authors have attempted to apply supervised contrastive loss~\cite{chen2023writing} as part of a style change detection system. These works could benefit from a more efficient large-scale approach, utilizing larger batches with significantly more negative and positive examples while applying them to more diverse datasets. Large-scale pretraining has proven to be beneficial in works like CLIP and SimSCE; therefore, it is reasonable to expect similar benefits in the context of authorship understanding.

In our approach, we combine pretrained transformers, which already possess some stylometric capabilities, to reduce training time, while implementing a large-scale approach to contrastive learning in order to construct universal author writing style representations.

\section{Authorship embeddings}\label{sec:methods}
Authorship embeddings are representations of the abstract features of speech like punctuation, rhythm and registry, they encompass everything about the author's style of writing. These patterns can be either intentional in the case of professional writers or unintended such as with social media users. Independently of intentionality these patterns can be measured ~\cite{neal2017surveying} by an expert, who could hand-craft numerical features from any given text. As this is a time-consuming task we aim to automate this feature extraction process to generate unique embeddings from each author. In an ideal scenario every author would be unequivocally represented by a single embedding, independently of the text, but this is impossible as style can vary from one text to another, for instance a poet could write an essay or a twitter user could write a fan-fiction story. Thus we can expect for two authorship embeddings belonging to the same author to be different but, at least, very similar. This is comparable to the semantic embedding generated by encoder-only transformers where for two phrases with the exact same meaning, the embeddings are not identical but similar enough to identify that they belong to the same concept. Generating this type of embedding has been studied previously in the literature~\cite{Huertas-Tato2022Jul, wegmann-etal-2022-author} thus we improve upon previous work applying our novel methodology.

\textbf{Problem definition:} In short, authorship embeddings are defined as a numerical representation of the writing style of an author and our aim is to numerically approximate their features of speech. To this end, we require documents organized by author thus, let $A = \{A^{(1)}, A^{(2)}, ..., A^{(n)}\}$ be the set of authors with each author containing an authored document set such as $A^{(i)} =  \{d^{(i)}_{1}, d^{(i)}_{2}, ..., d^{(i)}_{m}\}$ where $n$ is the number of authors and $m$ is the number of documents. For each document we want to generate an embedding such as $e^{(i)}_j = f(d^{(i)}_{j})$ to have a numerical representation of a text, forming an embedding set $E = \{E^{(1)}, E^{(2)}, ..., E^{(n)}\}$ where $E^{(i)}= \{e^{(i)}_1, e^{(i)}_2, ..., e^{(i)}_m\}$. Our objective is to get an encoder function that given two documents written by the same author, their cosine similarity is $cos(e^{(i)}_a, e^{(i)}_b) = 1$, where any two embeddings from the same source have maximum similarity; while embeddings from different sources have minimal similarity $cos(e^{(i)}_a, e^{(j)}_b) = 0$. Thus we want to approximate the following Eq.~\ref{eq:ideal}

\begin{equation}\label{eq:ideal}
    cos(f(d^{(i)}_{j}), f(d^{(i)}_{j}))=
    \begin{cases}
      1 & \text{if}\ i=j \\
      0 & \text{otherwise}
    \end{cases}
\end{equation}

Using this conceptual framework we infer that the best way to approximate a similarity function like this is using contrastive learning. In particular Supervised Contrastive Training~\cite{Khosla2020Apr} has properties that we can use to develop our desired encoder function $f$. 

%Contrastive learning section

\subsection{Supervised Contrastive Pretraining for Authorship Representations} \label{sec:scp} %% Explainable figure
\begin{figure}[h]
\centering
\includegraphics[width=\textwidth]{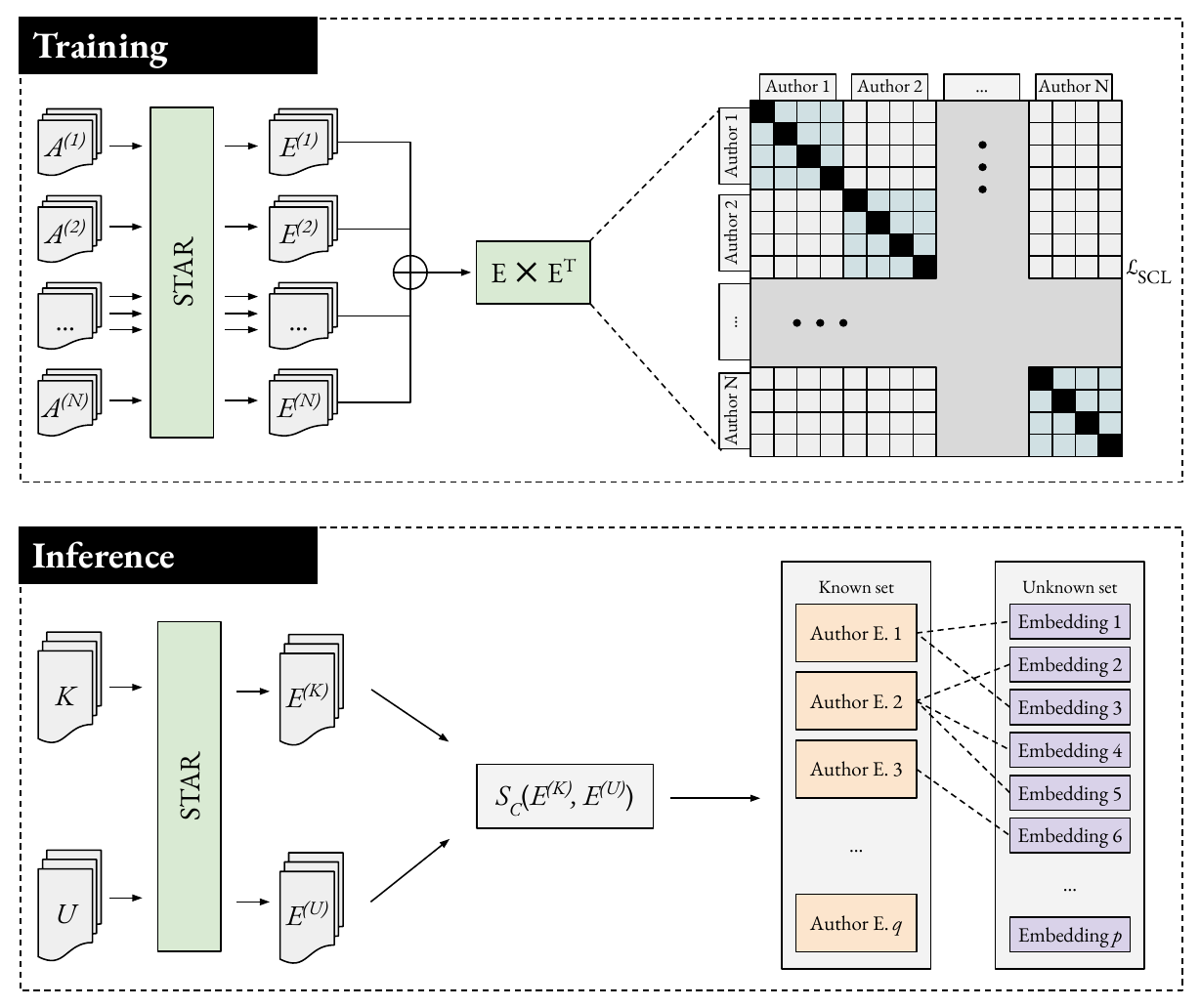}
\caption{Summary of STAR training and inference.} \label{fig:summary}
\end{figure}

A summary of the used method is presented in Fig.~\ref{fig:summary}. For learning document author embeddings we use supervised contrastive loss (or SupCon loss for short). Each document of an author $A$ is considered a view for the purpose of SupCon, thus picking $k$ authors and $l$ random documents from these authors results in a multi-viewed batch of authored documents. We avoid repeating the same author twice within the batch, so it has $l$ positive documents for each author $i$ with indices $P(i)$, and $(k-1)*l$ negative documents with indices $A(i)$. For each document we want to generate an embedding such as $e_j = f(d_{j})$.

We use the SupCon loss term as presented in Eq.~\ref{eq:supcon}:

\begin{equation}\label{eq:supcon}
\mathcal{L}_{SCL}(e) = \sum_{i \in I} \frac{-1}{|P(i)|} \sum_{p \in P(i)} \log\frac{\exp(e_i \cdot e_p/ \tau)}{\sum_{a \in A(i)} \exp(e_i \cdot e_a / \tau)}
\end{equation}

where $|P(i)|$ is the cardinality and $\tau$ is a trainable temperature parameter. 

With this loss term similar positive examples are rewarded while dissimilar negative examples are also rewarded; related documents are grouped together and will present greater cosine similarity with same-author text than any other text. This way we force the transformer to learn authorship representations, which requires to contain defining common features of style to minimize loss. 

\textbf{Inference example on attribution:} Performing inference requires a document set $K$ from the known authors, and a document set from anonymous sources $U$. These to sets of documents are processed through STAR extracting $E^{(K)}$ and $E^{(U)}$ embeddings. Each unknown document $j$ is matched to the author embedding with highest cosine similarity using $\max_{i} S_c(E^{(K)}_i, E^{(U)}_j)$

We further explore how to use the model in the experimentation, applying our methods to attribution and fine-tuning a minimal model for verification.

\section{Experimental setup}\label{sec:setup}
The following section describes the experimental setup for the experiments in section \ref{sec:results}. Our method requires pre-training thus is very reliant on the initial data usage, which we describe. This includes our domain data for training, as well as the out-of-domain data for validation.

\subsection{Training data handling}
We aim to learn a high variety of textual stylistic features, therefore data requires to match this variety to recognize new, unseen authors successfully. Other key to this technique is how to handle the data, over-representation of a single style can be harmful but some datasets have limited amount of authors. 

Constructing a batch is performed by sampling $l$ texts from $n$ different authors authors. Generalization of the network is higher as the batch size $b=l \cdot n$, but in-batch representation affects the capacity of the model as reviewed in previous work, therefore each batch is balanced equally ensuring fair representation for all datasets. Authors from underrepresented datasets are upsampled to match larger datasets. Sampling texts from an author is performed at random, this mitigates the upsampling overrepresentation of some authors in the dataset. We do not consider documents as datapoints instead we consider each author to be a point in data, therefore the number of texts per authors does not lead to imbalance.

Texts written by the same author may be longer than the maximum determined length, this is common on datasets containing books, therefore all texts are split in equal sized chunks without overlap. Only authors with more than 16 texts are introduced to the training dataset. If a text has been written by someone anonymous or by several authors, the text is not included in the dataset either, independently of the number of texts found.

Additionally some datasets have received special consideration and preprocessing described as follows:

\textbf{Standardized Gutenberg~\cite{gerlach2020standardized}}: The Gutenberg Project is an open repository of literary works from authors whose works have entered the public domain. The writing styles range from novels, to poetry and essays although it presents imbalance in the representation. Works are divided into books, each book contains identifying information within the first and last text chunks, which are removed from the dataset. After preprocessing, we have extracted 1270 authors with, on average, 542 documents. There is a total of $6.89 \cdot 10^5$ documents in this dataset.

\textbf{Blog authorship~\cite{schler2006effects}}: The blog profiling dataset was originally constructed to study the age, gender and other personal features from textual content. We use it for attribution with our methods. It contains text ranging from personal opinion to short stories or fan-fiction. No additional preprocessing was required for this dataset. We found 4837 eligible authors, with 60 documents per author, for a total of $2.9 \cdot 10^5$.

\textbf{Twitter users~\cite{cheng2010you}}: Using the geolocation dataset we retrieve 10000 tweets (or all, whichever is lower) from all users in the original dataset. We clean each twit removing user information and links, replaced by special tokens \textit{<h>} and \textit{<u>} for hyperlinks and usernames respectively. To make training more efficient we join tweets with a triple linespace to fill up the maximum 512 token length. This dataset contains stylistically varied twits, including emoji usage, slang, irony, shorthand and other informal registry text. At training time, there were 52465 accounts, with 57 documents available on average. A total of $3.02 \cdot 10^6$ documents are retrieved from this dataset.

\textbf{Reddit TLDR~\cite{volske-etal-2017-tl}}: The Reddit TLDR dataset contains texts from reddit users posted in different forums explaining several topics and ranging from informal divulgation to gossip. We split this dataset for testing. Both datasets require no further preprocessing, the train dataset has 14548 authors with 36 documents on average and the testing dataset has 1616 authors with 29 documents on average. The training dataset contains $2.9 \cdot 10^5$ documents and the testing dataset $2.9 \cdot 10^5$.

\textbf{\textit{Full dataset}}: In total, we balance the datasets so each represents 1/4th of the authors. The final dataset contains 73466 unique authors with $4.5 \cdot 10^6$ texts to train with.

\subsection{Validation method}
We extract experimental results with three different approaches. First we explore the PAN authorship attribution common tasks, to test the zero-shot capabilities of the model. The second section explores the PAN style change detection, to showcase the capabilities of fine-tuning the STAR model. Finally we showcase the application of the model in a practical domain using the reddit TL;DR testing dataset.

On all experiments we focus on comparing our new method STAR against PART and RoBERTa, and for zero-shot methods we also include Style-embeddings and DeBERTa-v3. PART is the previous version of the model without the described upgrades, on the other hand RoBERTa is the backbone of PART and the warm startup weights of STAR. We do not report scores for the competitions due to them being highly engineered towards each challenge, making the comparison both unfair and biased.

\textit{PAN Attribution}: For attribution we use the described method in Section~\ref{sec:scp}, we first split each candidate document in chunks of 512; compute the embeddings of candidates and unknown documents; perform the pairwise cosine similarity and assign each unknown document to its most similar candidate. We measure F1-Score and Accuracy according to the competitions main metrics. We explore all available challenges on this category.%~\cite{argamon2011overview, juola2012overview, kestemont2018overview, kestemont2019overview}.

\begin{figure}[h] 
\centering
\includegraphics[width=\textwidth]{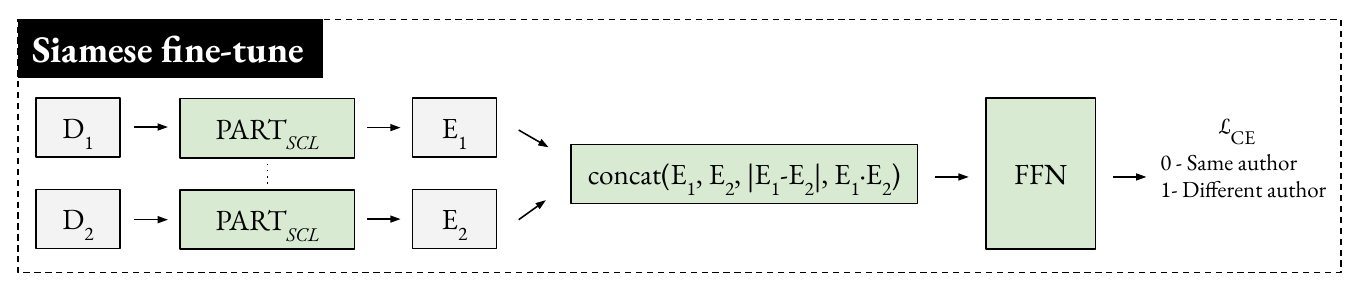}
\caption{Siamese architecture for fine tuning.}\label{fig:siam}
\end{figure}

\textit{PAN Authorship verification}: For verification we design a simple siamese topology to classify whether two texts belong to the same author or not. We extract both embeddings from the frozen models and concatenate each embedding along with their absolute difference and their product, which is classified by a dense layer and a output neuron (a feed forward network). The siamese architecture is presented in Figure~\ref{fig:siam}. As this method requires fine-tuning we present results from RoBERTa and PART alongside STAR, RoBERTa is the main baseline and PART is the second best model in earlier evaluations. We also present the best results of each competition due to the fine-tuning requirement allowing our model to compete fairly in a non zero-shot way.

\textit{PAN Author Clustering}: For verification we design a simple siamese topology to classify whether two texts belong to the same author or not. We extract both embeddings from the frozen models and concatenate each embedding along with their absolute difference and their product, which is classified by a dense layer and a output neuron (a feed forward network). The siamese architecture is presented in Figure~\ref{fig:siam}.

\textit{Reddit application}: The testing data of reddit is treated in a similar fashion to attribution, however as there is no defined candidate and unknown set, we randomly sample both. We also have interest in measuring the capabilities of the embeddings when they are combined. We retrieve $N_p=[1, 2, 4, 8]$ examples as each candidate document and try to infer which unknown document from a set of $N_n=[10, 20, 50, 100, 250, 500, all]$ author candidates with the same number. 

\subsection{Implementation details}
An encoder is required for $e_j = f(d_{j})$ so we use a encoder-only transformer model. Warm-starting a model is more efficient time-wise for pre-training, thus we use the public RoBERTa~\cite{liu2019roberta} large checkpoint. The RoBERTa transformer outputs a weighted average of word embeddings, to form a sentence embedding representative of the input content, this sentence embedding is processed by a final dense layer with dimension $d$ containing the final authorship embedding, with no activation whatsoever, we compute embeddings in the same way. Hyper-parameters used in training are shown in Table~\ref{tab:hyper}.

\begin{table}[]\label{tab:hyper}
\centering
\begin{tabular}{@{}rl@{}} 
\toprule
\multicolumn{1}{l}{} & \textbf{Training hyperparameters} \\ \midrule
Batch size           & \textit{b=16384 $\rightarrow$ l=1024, k=16}    \\
Max sequence length  & \textit{512}                      \\
Base model           & \textit{RoBERTa large}            \\
Training steps       & \textit{3000}                     \\
Schedule             & \textit{Warmup with linear decay} \\
Warmup steps          & \textit{180}                      \\
Optimizer            & \textit{AdamW}                    \\
Learning Rate        & \textit{1e-2}                     \\
Weight decay         & \textit{1e-4}                     \\ \bottomrule
\end{tabular}
\caption{Hyper-parameters used for pretraining the authorship model.}
\end{table}

Training and validation has been performed on a Titan V GPU over the course of a month given the described hyperparameters with the huggingface \textit{roberta-large}\footnote{Huggingface \& RoBERTA$_{LARGE}$: \hyperlink{https://huggingface.co/roberta-large}{ https://huggingface.co/roberta-large}} checkpoint for the warm starting of training. We have used the PyTorch \footnote{Pytorch: \href{https://pytorch.org/}{pytorch.org}} backend with the Lightning API \footnote{Pytorch Lightning: \href{https://lightning.ai/docs/pytorch/stable/}{lightning.ai/docs/pytorch/stable/}} to perform and log the experiments. We have uploaded the model to huggingface as aida-upm/star \footnote{STAR in huggingface: \hyperlink{huggingface.co/aida-upm/star}{aida-upm/star}} for public usage. 

\section{Experimental results} \label{sec:results}
Following our validation setup described we present the experimental results below.

\subsection{Authorship attribution}
Results for each competition are shown in tables~\ref{tab:attribution_pan_global} and~\ref{tab:attribution_pan_specific}. On Table~\ref{tab:attribution_pan_global} the global results are explored, computed as the average among related problems in the competition. On average, the new technique dominates earlier versions and baselines, specially in PAN19. PAN19 is particularly close to the training domain due to it containing fanfiction which has style overlap with the blog dataset and the literary works of the gutenberg corpus. In practice, the PAN19 dataset achieves very high accuracy in english outclassing the best reported method in the competition by 4.5\% f1-score.

\begin{table}[]

\centering
\resizebox{\textwidth}{!}{%
\begin{tabular}{@{}l|lllll|lllll@{}}
\toprule
               & \multicolumn{5}{c|}{Accuracy}                       & \multicolumn{5}{c}{F1-Score}                        \\
Challenge &
  \textit{STAR} &
  \textit{PART} &
  \textit{Style-Embedding} &
  \textit{RoBERTa} &
  \textit{DeBERTa-v3} &
  \textit{STAR} &
  \textit{PART} &
  \textit{Style-Embedding} &
  \textit{RoBERTa} &
  \textit{DeBERTa-v3} \\ \midrule
PAN-11 - \textit{Large} & \textbf{0.3985} & 0.3754 & 0.2508 & 0.2869 & 0.1300 & \textbf{0.2809} & 0.2618 & 0.1647 & 0.1882 & 0.0815 \\
PAN-12         & \textbf{0.8730} & 0.8413 & 0.6617 & 0.6915 & 0.5585 & \textbf{0.8534} & 0.8216 & 0.6020 & 0.6354 & 0.5142 \\
PAN-18         & \textbf{0.6945} & 0.6920 & 0.5677 & 0.5635 & 0.3499 & \textbf{0.6206} & 0.5942 & 0.4733 & 0.4309 & 0.3210 \\
PAN-19         & \textbf{0.8038} & 0.6169 & 0.6214 & 0.5162 & 0.4070 & \textbf{0.7100} & 0.5464 & 0.4986 & 0.4450 & 0.3081 \\ \bottomrule
\end{tabular}%
}
\caption{Zero-shot results for PAN-CLEF attribution tasks averaged. \textit{Best} column indicates first place for the competition. The best scores for zero-shot methods marked in bold.} \label{tab:attribution_pan_global}
\end{table}

It is to be noted that he PAN11 challenge is derived from the Enron Mail dataset~\cite{shetty2004enron} which, as shown in previous work, was troubling for PART; our new methodology and training result in much higher accuracies and f1-scores. Despite the improvement we still score 2nd in the competition, off by 4\% from the best method. 

Finally, PAN 12 results in close scores to the best model, averaging to 85.4\%, behind by 0.5\%. PAN18 results are behind the best model in the cometition by 14\%. Summarizing, STAR outclasses previous zero-shot approximations to attribution on average, but still lags behind when not properly tuned to the specific domain. However, when the domain is tangentially close to our training styles, metrics improve by a large margin.

A breakdown of the problems is shown in Table~\ref{tab:attribution_pan_specific}. Here we see how the model STAR compares to previous work. STAR frequently outclasses other embeddings but there are some specific problems in the PAN challenge that are better solved by PART. There is no apparent pattern in this regard. Another observation is that the difficulty of the problem is not indicative of higher or lower accuracy, as exemplified by PAN18 - 10 authors and PAN18 - 5 authors. The first should be harder to solve but achieves the highest accuracy and F1-score, while the second has worse metrics despite being significantly easier. This phenomenon repeats on the PAN19 challenge too.

Across challenges we observe the underperforming of a vanilla RoBERTa large model, lagging behind both pretrained authorship models at almost every available problem. IN summary, we observe that STAR also works at a finer level and is generally better on most problems, including without specific training, making it very a very powerful feature extractor.

\begin{table}[]
\centering
\resizebox{\textwidth}{!}{%
\begin{tabular}{@{}l|lllll|lllll@{}}
\toprule
                   & \multicolumn{5}{c|}{Accuracy}                                         & \multicolumn{5}{c}{F1-Score}                                          \\
Challenge & \textit{STAR} & \textit{PART} & \textit{Style-Embedding} & \textit{RoBERTa} & \textit{DeBERTa-v3} & \textit{STAR} & \textit{PART} &  \textit{Style-Embedding} & \textit{RoBERTa} & \textit{DeBERTa-v3} \\ \midrule
PAN11 - \textit{Large}      & \textbf{0.3985} & 0.3754          & 0.2508          & 0.2869 & 0.1300 & \textbf{0.2809} & 0.2618          & 0.1647          & 0.1882 & 0.0815 \\
PAN12 - \textit{A}          & 0.8333          & 0.8333          & \textbf{1.0000} & 0.5000 & 0.6667 & 0.8222          & 0.8222          & \textbf{1.0000} & 0.4127 & 0.6556 \\
PAN12 - \textit{B}          & 0.8333          & \textbf{1.0000} & 0.6667          & 0.6667 & 0.6667 & 0.8222          & \textbf{1.0000} & 0.5556          & 0.6556 & 0.6556 \\
PAN12 - \textit{C}          & \textbf{1.0000} & 1.0000          & 0.6250          & 1.0000 & 0.6250 & \textbf{1.0000} & 1.0000          & 0.5833          & 1.0000 & 0.5833 \\
PAN12 - \textit{D}          & \textbf{1.0000} & 1.0000          & 0.7500          & 0.6250 & 0.2500 & \textbf{1.0000} & 1.0000          & 0.6875          & 0.5417 & 0.1667 \\
PAN12 - \textit{I}          & \textbf{0.8571} & 0.5714          & 0.5000          & 0.6429 & 0.5000 & \textbf{0.8095} & 0.5357          & 0.4167          & 0.5357 & 0.4524 \\
PAN12 - \textit{J} &  \textbf{0.7143} & 0.6429 & 0.4286 & \textbf{0.7143} & 0.6429 & \textbf{0.6667} & 0.5714 & 0.3690 & \textbf{0.6667} & 0.5714 \\
PAN18 - \textit{20 authors} & \textbf{0.6582} & 0.6203          & 0.4937          & 0.4684 & 0.2785 & \textbf{0.6566} & 0.5549          & 0.4131          & 0.3359 & 0.2308 \\
PAN18 - \textit{15 authors} & 0.5946          & \textbf{0.6351} & 0.5270          & 0.4730 & 0.2838 & \textbf{0.5788} & 0.5220          & 0.4288          & 0.3186 & 0.2323 \\
PAN18 - \textit{10 authors} & \textbf{0.9000} & 0.8250          & 0.6250          & 0.6250 & 0.4000 & \textbf{0.7557} & 0.7544          & 0.5029          & 0.5617 & 0.3482 \\
PAN18 - \textit{5 authors}  & 0.6250          & \textbf{0.6875} & 0.6250          & 0.6875 & 0.4375 & 0.4914          & 0.5455          & \textbf{0.5485} & 0.5073 & 0.4727 \\
PAN19 - \textit{r=100}\%    & \textbf{0.9252} & 0.8504          & 0.8248          & 0.7350 & 0.5962 & \textbf{0.8071} & 0.7166          & 0.6843          & 0.5730 & 0.4146 \\
PAN19 - \textit{r=80}\%     & \textbf{0.6327} & 0.4898          & 0.5102          & 0.4694 & 0.2551 & \textbf{0.5263} & 0.3986          & 0.4702          & 0.4036 & 0.2493 \\
PAN19 - \textit{r=60}\%     & \textbf{0.6742} & 0.5379          & 0.4924          & 0.5000 & 0.2803 & \textbf{0.6564} & 0.4714          & 0.4729          & 0.4665 & 0.2701 \\
PAN19 -\textit{ r=40}\%     & \textbf{0.9079} & 0.4868          & 0.8553          & 0.1645 & 0.4868 & \textbf{0.7877} & 0.6035          & 0.4327          & 0.2753 & 0.3359 \\
PAN19 - \textit{r=20}\%     & \textbf{0.8788} & 0.7197          & 0.4242          & 0.7121 & 0.4167 & \textbf{0.7723} & 0.5420          & 0.4327          & 0.5067 & 0.2708 \\ \bottomrule
\end{tabular}%
}
\caption{Zero-shot results for all PAN-CLEF attribution tasks. Accuracy and F1-Score are reported, best results reported in bold for each metric.}\label{tab:attribution_pan_specific}

\end{table}

\subsection{Authorship verification}
Table \ref{tab:verification} presents the comparison results among PART, our novel STAR approach, RoBERTa, and the top performer from various PAN author style change detection competitions. As an aside, we only compare against PART, which is second best in most cases of the previous evaluation and RoBERTa, which is the main baseline for our model, we exclude others
Our proposed method involves constructing a Siamese architecture (see Figure \ref{fig:siam}), where only a linear classification layer with 2 outputs is trained, leaving the remaining layers of the architecture unaltered. As of the date of this article, the results of the PART 2023 competition have not been published.

Notably, STAR outperforms PART in all cases and demonstrates significant improvements over Roberta. Even when compared to methods achieving the best results in the competition, our approach remains highly competitive. Despite not being a fully trained architecture on the entire dataset, it consistently approaches the performance of models trained with complete data. These results suggest that STAR possesses substantial potential for zero-shot classification tasks.

\begin{table}
\centering
\begin{tabular}{l|llll}
\toprule
Challenge & \textit{STAR} & \textit{PART}  & \textit{RoBERTa} & \textbf{Best result} \\ 
\midrule
PAN 2023 - \textit{Dataset 1}  & \textbf{0.9224} & 0.8669 & 0.8047 & - \\ 
PAN 2023 - \textit{Dataset 2}  & \textbf{0.7562} & 0.7439 & 0.4133 & - \\ 
PAN 2023 - \textit{Dataset 3} & \textbf{0.6384} & 0.6221 & 0.3615 & - \\ 
PAN 2022 - \textit{Task 2 Dataset 1} & \textbf{0.8370} & 0.7951 & 0.7951 & 0.7070 \\ 
PAN 2022 - \textit{Task 2 Dataset 2} & \textbf{0.7364} & 0.6932 & 0.4802 & 0.7070 \\ 
PAN 2022 - \textit{Task 2 Dataset 3} & \textbf{0.6312} & 0.6227 & 0.4013 & 0.7070 \\ 
PAN 2021 - \textit{Task 2} & \textbf{0.7043} & 0.6942 & 0.4013 & 0.7510 \\ 
PAN 2020 - \textit{Task 2 narrow} & \textbf{0.8793} & 0.8412 & 0.8271 & 0.8567 \\ 
PAN 2020 - \textit{Task 2 wide} & \textbf{0.8532} & 0.8238 & 0.7541 & 0.8567 \\ 
\bottomrule
\end{tabular}\label{tab:verification}
\caption{Results for 2020-2023 PAN-CLEF competition on writing style change detection for PART, $STAR$, RoBERTa and the best result obtained in the conference. It must be noted that PART, $STAR$ and RoBERTa columns show results where only one final classification layer is trained, while the rest of layers are frozen. The best result from the competition consider the best method, which usually includes feature engineering or training a whole architecture. F1-score is reported for each experiment.}
\end{table}

\subsection{Authorship clustering}
Table \ref{tab:clusteringsum} presents the results for both clustering challenges. STAR presents outstanding results in both challenges at zero-shot on the B cubed F1 score performance metric. Star achieves a 1.47\% improvement upon DeBERTa-v3 in the PAN16 challenge, as well as a 11.01\% increase compared to Style-Embedding on PAN17.

In this challenge PART is behind most other models, underperforming on this out-of-domain task. Clustering proves to be a difficult task without previous domain knowledge, other approaches may finetune the model to the data. Surprisingly the second best performing model is RoBERTa for the PAN17 challenge, which may confirm that pre-trained models already have understanding over structure.

\begin{table}[]
\resizebox{\textwidth}{!}{%
\begin{tabular}{l|rrrrr}
\hline
      & \multicolumn{5}{c}{F1-Score}                        \\
Challenge & \textit{STAR} & \textit{PART} & \textit{Style-Embedding} & \textit{RoBERTa} & \textit{DeBERTa-v3} \\ \hline
PAN16 & \textbf{0.8340} & 0.8173 & 0.8194 & 0.8163 & 0.8219 \\
PAN17 & \textbf{0.5763} & 0.4452 & 0.4932 & 0.5191 & 0.4910 \\ \hline
\end{tabular}%
}
\caption{Average of clustering challenge results at zero-shot.}
\label{tab:clusteringsum}
\end{table}

We confirm that STAR makes for the best clustering in Table \ref{tab:clustering}, where we show the model topping performance on 18/26 problems, acting in many cases as a close second. We have no more meaningful observations from this evaluation other that, on average, STAR is the best choice for most problems and that the baselines have no meaningful differences other than being worse than STAR.

\begin{table}[]
\resizebox{\textwidth}{!}{%
\begin{tabular}{l|lllll}
\hline
                     & \multicolumn{5}{l}{F1-Score}                                                            \\
Challenge          & \textit{STAR}   & \textit{PART}   & \textit{Style-Embedding} & \textit{RoBERTa} & \textit{DeBERTa-v3} \\ \hline
PAN16 - \textit{Problem 1}  & \textbf{0.8333} & \textbf{0.8333} & \textbf{0.8333}          & \textbf{0.8333}  & \textbf{0.8333}     \\
PAN16 - \textit{Problem 2}    & \textbf{0.9552} & \textbf{0.9552} & \textbf{0.9552} & 0.9214          & \textbf{0.9552} \\
PAN16 - \textit{Problem 3}    & 0.9417          & \textbf{0.9474} & \textbf{0.9474} & \textbf{0.9474} & \textbf{0.9474} \\
PAN16 - \textit{Problem 4}    & \textbf{0.7524} & 0.6667          & 0.6667          & 0.6913          & 0.6667          \\
PAN16 - \textit{Problem 5}    & \textbf{0.8463} & 0.8316          & 0.8342          & 0.8316          & 0.8406          \\
PAN16 - \textit{Problem 6}    & 0.6747          & 0.6698          & 0.6794          & 0.6731          & \textbf{0.6885} \\
PAN17 - \textit{Problem 1}    & \textbf{0.6953} & 0.6207          & 0.5286          & 0.5851          & 0.6499          \\
PAN17 - \textit{Problem 2}    & 0.4689          & \textbf{0.5630} & 0.5062          & 0.5193          & 0.3815          \\
PAN17 - \textit{Problem 3}    & 0.5353          & 0.4228          & 0.5308          & 0.5060          & \textbf{0.6440} \\
PAN17- \textit{Problem 4}     & \textbf{0.5380} & 0.4000          & 0.4009          & 0.5147          & 0.3871          \\
PAN17- \textit{Problem 5}     & \textbf{0.5075} & 0.3562          & 0.4997          & 0.4713          & 0.4724          \\
PAN17- \textit{Problem 6}     & \textbf{0.6696} & 0.4205          & 0.4154          & 0.3913          & 0.4000          \\
PAN17 - \textit{Problem 7}    & \textbf{0.5124} & 0.4615          & 0.4953          & 0.4792          & 0.3897          \\
PAN17 - \textit{Problem 8}    & \textbf{0.6438} & 0.5693          & 0.5044          & 0.5622          & 0.5000          \\
PAN17 - \textit{Problem 9}    & 0.5990          & 0.4518          & 0.5226          & 0.6020 &           \\
PAN17 - \textit{Problem 10} & \textbf{0.4000} & \textbf{0.4000} & \textbf{0.4000}          & \textbf{0.4000}  & \textbf{0.4000}     \\
PAN17 - \textit{Problem 11}   & \textbf{0.5888} & 0.5185          & 0.5826          & 0.5658          & 0.5377          \\
PAN17 - \textit{Problem 12}   & 0.4520          & \textbf{0.5577} & 0.5326          & 0.4192          & 0.5073          \\
PAN17 - \textit{Problem 13}   & \textbf{0.6399} & 0.4000          & 0.5901          & 0.6033          & 0.4579          \\
PAN17 - \textit{Problem 14}   & \textbf{0.5994} & 0.5401          & 0.5201          & 0.5585          & 0.5115          \\
PAN17 - \textit{Problem 15}   & 0.4426          & 0.3333          & \textbf{0.4986} & 0.4146          & 0.3596          \\
PAN17 - \textit{Problem 16}   & \textbf{0.5467} & 0.3304          & 0.3608          & 0.5137          & 0.4392          \\
PAN17 - \textit{Problem 17}   & \textbf{0.7677} & 0.3439          & 0.4238          & 0.5765          & 0.3618          \\
PAN17 - \textit{Problem 18}   & \textbf{0.8107} & 0.2725          & 0.3455          & 0.4495          & 0.6046          \\
PAN17 - \textit{Problem 19}   & 0.5074          & 0.3304          & 0.5854          & \textbf{0.6287} & 0.5813          \\
PAN17 - \textit{Problem 20} & 0.6000          & 0.6107          & \textbf{0.6207} & \textbf{0.6207} & \textbf{0.6207} \\ \hline
\end{tabular}%
}\caption{Results for clustering PAN challenges at zero-shot, all problems presented. The best result for each problem has been bolded.}

\label{tab:clustering}

\end{table} 

\subsection{Application on Reddit}
Finally we share our results on a test partition on the reddit dataset. We test over 1616 authors unseen in training to inspect them at a known domain of social networks. In Tables~\ref{tab:reddit attribution} and~\ref{tab:reddit_attribution_ex} we show the Accuracy and Top-5 Accuracy of our method compared to the baselines.

\begin{table}[]
\resizebox{\textwidth}{!}{%
\begin{tabular}{rr|lllllll@{}}
\toprule
Method &
  Positive examples &
  $N_n=10$ &
  $N_n=20$ &
  $N_n=50$ &
  $N_n=100$ &
  $N_n=250$ &
  $N_n=500$ &
  $N_n=1616$ \\ \midrule
\textit{STAR} &
  $N_p=1$ &
  \textbf{76.05\% ± 13.83} &
  \textbf{68.58\% ± 9.46} &
  \textbf{54.29\% ± 6.99} &
  \textbf{46.05\% ± 4.51} &
  \textbf{37.05\% ± 2.89} &
  \textbf{30.37\% ± 1.61} &
  \textbf{21.39\% ± 0.82} \\
 &
  $N_p=2$ &
  \textbf{91.45\% ± 7.85} &
  \textbf{84.40\% ± 8.24} &
  \textbf{77.44\% ± 4.94} &
  \textbf{70.68\% ± 4.26} &
  \textbf{60.67\% ± 2.61} &
  \textbf{53.08\% ± 1.87} &
  \textbf{41.35\% ± 0.86} \\
 &
  $N_p=4$ &
  \textbf{97.65\% ± 3.97} &
  \textbf{95.90\% ± 3.63} &
  \textbf{91.97\% ± 2.98} &
  \textbf{86.92\% ± 2.56} &
  \textbf{80.74\% ± 1.96} &
  \textbf{74.83\% ± 1.61} &
  \textbf{63.86\% ± 0.74} \\
 &
  $N_p=8$ &
  \textbf{99.30\% ± 2.12} &
  \textbf{98.47\% ± 2.12} &
  \textbf{97.11\% ± 1.81} &
  \textbf{95.39\% ± 1.82} &
  \textbf{91.65\% ± 1.38} &
  \textbf{88.06\% ± 1.08} &
  \textbf{80.42\% ± 0.49} \\ \midrule
\textit{PART} &
  $N_p=1$ &
  52.40\% ± 15.22 &
  43.35\% ± 9.53 &
  30.12\% ± 5.89 &
  24.32\% ± 4.03 &
  17.24\% ± 2.21 &
  13.53\% ± 1.34 &
  8.55\% ± 0.60 \\
 &
  $N_p=2$ &
  72.00\% ± 10.72 &
  62.82\% ± 9.41 &
  50.56\% ± 6.32 &
  43.40\% ± 4.08 &
  33.18\% ± 2.40 &
  26.73\% ± 1.51 &
  18.64\% ± 0.67 \\
 &
  $N_p=4$ &
  87.50\% ± 9.39 &
  81.02\% ± 8.04 &
  70.96\% ± 5.89 &
  63.78\% ± 4.16 &
  52.95\% ± 2.78 &
  45.59\% ± 1.50 &
  34.61\% ± 0.79 \\
 &
  $N_p=8$ &
  94.85\% ± 4.97 &
  90.95\% ± 5.29 &
  85.11\% ± 4.55 &
  78.92\% ± 3.07 &
  70.64\% ± 2.28 &
  63.61\% ± 1.60 &
  51.67\% ± 0.64 \\ \midrule
\textit{Style-Embeddings} &
  $N_p=1$ &
  36.10\% ± 14.41 &
  24.63\% ± 9.03 &
  13.72\% ± 3.81 &
  9.26\% ± 2.33 &
  5.55\% ± 1.22 &
  3.66\% ± 0.78 &
  1.73\% ± 0.25 \\
 &
  $N_p=2$ &
  43.15\% ± 11.39 &
  30.33\% ± 7.90 &
  18.92\% ± 4.46 &
  12.83\% ± 2.62 &
  7.53\% ± 1.23 &
  5.08\% ± 0.66 &
  2.54\% ± 0.26 \\
 &
  $N_p=4$ &
  47.45\% ± 8.50 &
  35.23\% ± 7.22 &
  23.57\% ± 4.07 &
  16.75\% ± 2.60 &
  10.93\% ± 1.22 &
  7.20\% ± 0.72 &
  3.74\% ± 0.27 \\
 &
  $N_p=8$ &
  52.05\% ± 9.90 &
  40.70\% ± 7.65 &
  27.26\% ± 3.83 &
  20.01\% ± 2.43 &
  13.05\% ± 1.25 &
  9.09\% ± 0.66 &
  5.09\% ± 0.28 \\ \midrule
\textit{RoBERTa}&
  $N_p=1$ &
  25.50\% ± 12.05 &
  15.75\% ± 7.60 &
  10.47\% ± 3.82 &
  7.44\% ± 2.41 &
  5.09\% ± 1.17 &
  3.71\% ± 0.69 &
  2.31\% ± 0.33 \\
 &
  $N_p=2$ &
  28.50\% ± 9.96 &
  20.22\% ± 7.25 &
  13.16\% ± 3.37 &
  9.89\% ± 2.37 &
  6.21\% ± 1.22 &
  4.56\% ± 0.67 &
  2.78\% ± 0.29 \\
 &
  $N_p=4$ &
  32.95\% ± 11.02 &
  24.75\% ± 7.15 &
  17.14\% ± 3.99 &
  12.15\% ± 2.46 &
  8.08\% ± 1.32 &
  6.11\% ± 0.84 &
  3.75\% ± 0.32 \\
 &
  $N_p=8$ &
  37.95\% ± 10.39 &
  28.55\% ± 6.85 &
  20.62\% ± 4.73 &
  15.44\% ± 2.88 &
  10.73\% ± 1.32 &
  8.07\% ± 0.88 &
  5.12\% ± 0.37 \\ \midrule
\textit{DeBERTa-v3} &
  $N_p=1$ &
  17.15\% ± 10.54 &
  10.05\% ± 5.27 &
  4.79\% ± 2.78 &
  3.25\% ± 1.58 &
  1.63\% ± 0.74 &
  1.08\% ± 0.42 &
  0.55\% ± 0.15 \\
 &
  $N_p=2$ &
  21.10\% ± 9.10 &
  12.20\% ± 5.28 &
  6.42\% ± 2.47 &
  3.72\% ± 1.51 &
  1.99\% ± 0.63 &
  1.33\% ± 0.38 &
  0.67\% ± 0.15 \\
 &
  $N_p=4$ &
  23.55\% ± 9.09 &
  15.70\% ± 5.49 &
  8.42\% ± 2.51 &
  5.19\% ± 1.50 &
  3.10\% ± 0.72 &
  1.98\% ± 0.41 &
  1.00\% ± 0.16 \\
 &
  $N_p=8$ &
  27.30\% ± 8.79 &
  17.15\% ± 5.10 &
  10.38\% ± 2.83 &
  7.18\% ± 1.73 &
  4.09\% ± 0.79 &
  2.79\% ± 0.49 &
  1.52\% ± 0.18 \\ \bottomrule
\end{tabular}%
}
\caption{Accuracy for attribution in reddit, presented as the average of 100 trials. $N_P$ is the number of support positives, or documents per author, and $N_n$ is the number of authors. Best results bolded.}
\label{tab:reddit attribution}
\end{table}

STAR outclasses every model by large, both in the stability over the trials and the accuracy itself. We observe the degradation of results when adding more authors to the trial and removing positives from our support base. As it could be expected the best results can be achieved with maximum support ($N_p=8$) and minimum difficulty ($N_n=10$), which is unsurprising but accurate nonetheless. Even with 100 authors accuracy surpasses 95\% and 99.69\% on top-5 accuracy when STAR is used.
\begin{table}[]
\resizebox{\textwidth}{!}{%
\begin{tabular}{@{}rr|lllllll@{}}
\toprule
Method &
  Positive examples &
  $N_n=10$ &
  $N_n=20$ &
  $N_n=50$ &
  $N_n=100$ &
  $N_n=250$ &
  $N_n=500$ &
  $N_n=1616$ \\ \midrule
\textit{STAR} &
  $N_p=1$ &
  \textbf{97.25\% ± 4.60} &
  \textbf{91.92\% ± 5.68} &
  \textbf{81.40\% ± 5.62} &
  \textbf{72.49\% ± 4.04} &
  \textbf{60.48\% ± 3.26} &
  \textbf{51.48\% ± 1.89} &
  \textbf{37.99\% ± 0.91} \\
 &
  $N_p=2$ &
  \textbf{99.75\% ± 1.09} &
  \textbf{98.57\% ± 2.13} &
  \textbf{95.26\% ± 2.21} &
  \textbf{90.78\% ± 2.34} &
  \textbf{83.15\% ± 1.81} &
  \textbf{75.55\% ± 1.65} &
  \textbf{62.83\% ± 0.91} \\
 &
  $N_p=4$ &
  \textbf{100.00\% ± 0.00} &
  \textbf{99.85\% ± 0.69} &
  \textbf{99.31\% ± 0.97} &
  \textbf{97.92\% ± 1.29} &
  \textbf{95.14\% ± 1.02} &
  \textbf{91.54\% ± 1.09} &
  \textbf{83.24\% ± 0.56} \\
 &
  $N_p=8$ &
  \textbf{100.00\% ± 0.00} &
  \textbf{100.00\% ± 0.00} &
  \textbf{99.93\% ± 0.26} &
  \textbf{99.69\% ± 0.43} &
  \textbf{98.89\% ± 0.49} &
  \textbf{97.67\% ± 0.48} &
  \textbf{93.74\% ± 0.36} \\ \midrule
\textit{PART} &
  $N_p=1$ &
  88.10\% ± 8.99 &
  75.15\% ± 9.82 &
  57.25\% ± 6.13 &
  46.52\% ± 4.18 &
  34.14\% ± 2.76 &
  26.93\% ± 1.89 &
  17.36\% ± 0.78 \\
 &
  $N_p=2$ &
  95.85\% ± 5.05 &
  89.55\% ± 5.77 &
  78.72\% ± 5.59 &
  68.76\% ± 4.10 &
  55.76\% ± 2.69 &
  46.60\% ± 1.89 &
  33.69\% ± 0.82 \\
 &
  $N_p=4$ &
  99.35\% ± 2.08 &
  97.10\% ± 3.44 &
  92.10\% ± 2.95 &
  86.23\% ± 2.95 &
  76.53\% ± 2.16 &
  68.43\% ± 1.59 &
  54.71\% ± 0.81 \\
 &
  $N_p=8$ &
  99.90\% ± 0.70 &
  99.50\% ± 1.06 &
  97.84\% ± 1.72 &
  95.00\% ± 1.72 &
  89.70\% ± 1.70 &
  84.21\% ± 1.26 &
  72.96\% ± 0.56 \\ \midrule
\textit{Style-Embeddings} &
  $N_p=1$ &
  78.45\% ± 11.42 &
  60.48\% ± 9.14 &
  37.61\% ± 5.43 &
  26.61\% ± 3.94 &
  15.81\% ± 1.91 &
  10.35\% ± 1.18 &
  4.99\% ± 0.41 \\
 &
  $N_p=2$ &
  82.30\% ± 9.50 &
  66.82\% ± 7.11 &
  46.65\% ± 5.17 &
  33.27\% ± 3.53 &
  20.62\% ± 1.93 &
  14.15\% ± 1.24 &
  7.19\% ± 0.47 \\
 &
  $N_p=4$ &
  90.15\% ± 6.80 &
  72.55\% ± 7.19 &
  52.66\% ± 4.43 &
  39.70\% ± 3.10 &
  26.33\% ± 1.93 &
  18.58\% ± 1.12 &
  10.14\% ± 0.43 \\
 &
  $N_p=8$ &
  92.50\% ± 6.54 &
  78.87\% ± 5.75 &
  58.92\% ± 4.31 &
  44.15\% ± 3.42 &
  30.11\% ± 1.83 &
  22.01\% ± 1.03 &
  12.78\% ± 0.41 \\ \midrule
\textit{RoBERTa} &
  $N_p=1$ &
  64.25\% ± 12.01 &
  42.45\% ± 9.69 &
  26.00\% ± 5.29 &
  17.92\% ± 3.47 &
  11.52\% ± 1.70 &
  8.31\% ± 1.09 &
  4.87\% ± 0.48 \\
 &
  $N_p=2$ &
  69.35\% ± 9.63 &
  49.15\% ± 8.01 &
  30.70\% ± 4.94 &
  21.88\% ± 3.16 &
  14.17\% ± 1.68 &
  10.32\% ± 1.02 &
  6.06\% ± 0.43 \\
 &
  $N_p=4$ &
  74.05\% ± 9.07 &
  55.68\% ± 6.81 &
  37.27\% ± 4.61 &
  27.31\% ± 3.22 &
  18.15\% ± 1.77 &
  13.31\% ± 1.08 &
  7.98\% ± 0.43 \\
 &
  $N_p=8$ &
  77.50\% ± 8.11 &
  58.98\% ± 7.48 &
  42.56\% ± 5.24 &
  32.39\% ± 3.52 &
  22.47\% ± 1.84 &
  16.87\% ± 1.01 &
  10.58\% ± 0.47 \\ \midrule
\textit{DeBERTa-v3} &
  $N_p=1$ &
  60.20\% ± 12.41 &
  35.18\% ± 9.32 &
  17.61\% ± 4.48 &
  10.71\% ± 2.41 &
  5.25\% ± 1.33 &
  3.11\% ± 0.69 &
  1.42\% ± 0.22 \\
 &
  $N_p=2$ &
  64.45\% ± 9.54 &
  40.03\% ± 6.99 &
  21.64\% ± 3.95 &
  13.24\% ± 2.64 &
  6.75\% ± 1.36 &
  4.27\% ± 0.65 &
  1.90\% ± 0.23 \\
 &
  $N_p=4$ &
  67.55\% ± 8.59 &
  44.45\% ± 7.45 &
  25.91\% ± 3.92 &
  16.42\% ± 2.39 &
  9.43\% ± 1.29 &
  5.93\% ± 0.74 &
  2.81\% ± 0.25 \\
 &
  $N_p=8$ &
  70.05\% ± 8.08 &
  46.83\% ± 6.18 &
  28.52\% ± 3.54 &
  19.53\% ± 2.32 &
  11.20\% ± 1.20 &
  7.48\% ± 0.67 &
  3.85\% ± 0.22 \\ \bottomrule
\end{tabular}%
}
\caption{Top-5 accuracy for attribution in reddit, presented as the average of 100 trials. $N_P$ is the number of support positives, or documents per author, and $N_n$ is the number of authors. Best results bolded.}
\label{tab:reddit_attribution_ex}
\end{table}

As previously stated the other models simply can't compete, fir example the closes competitor in all categories is PART, which achieves 94.85\% with $N_p=8$ \& $N_n=10$, STAR achieves 99.3\% in the same scenario with less dispersion. Other cases are 52\% on Style-Embeddings and 37.95\% on RoBERTa.

The quality is abysmal with 1 positive, but STAR is still much more accurate than its peers, achieving 21.39\% accuracy and 37.99\% accuracy picking an athor from a $N_n=1616$ with only 1 document, which is impressive by itself and dwarfs PART's results (8.55\% and 17.36\% accuracy and top-5). Other embeddings achieve percentages below 2.5\%.

\subsection{Discussion}
In our evaluation we have demonstrably determined that the embeddings produced by our pre-trained model are highly competitive in challenges and social media networks. In attribution and verification we find STAR dominating all baselines safe from some very specific cases, clustering is harder, being unreliable in more cases, while being the best choice bar none in the reddit test set.

In brief comparison, for the attribution challenges we achieve top-1 performance in the 2012 challenge and 2019 challenge, while achieving less than fifth place on the other challenges, performing strictly zero-shot techniques unlike every other method in leaderboards which usually rely on specifically designed and fine-tuned ensembles. This repeats for the verification leaderboard. The clustering best results outperform our model in both examined competitions but the model STAR itself achieves competitive performance without retraining.

Vanilla transformers can hold up at zero-shot, while style transformers (PART and Style-Embeddings) can underperform due to the lack of generalization outside of their training domains. This has been inspected over the several challenges and the hierarchy usually presents as PART in second place and Style embeddings third, with RoBERTa and DeBERTa-v3 being both at the bottom of performance due to their lack of pretraining in style. 

\section{Conclusion} \label{sec:conclusions}
We have presented STAR, a stylistic transformer with authorship as a pre-training contrastive objective. To build STAR we use a large batch of positive and negative documents, applying supervised contrastive learning to each minibatch to learn the distance from one embedding to another. The key to this study is using vast amounts of heterogeneous authored data, in particular we use three social media domains: Twitter for micro-blogging and slang texts, Reddit TL.DR for semi-formal internet usage, Blogs for a mixed bag of long-form texts and narration, and finally Project Gutenberg to cover for poetry, history and professional prose.With our mixture of data we achieve high zero-shot scores in clustering, attribution and verification challenges with little or no fine-tuning, getting close to achieving universal authorship embeddings. 

We demonstrate this performance on several State-of-the-art challenges and problems, including a validation in a real-world setting, a reddit testing partition. First we perform attribution, deciding which text belongs to which author given a support set of texts; we perform clustering where a group of texts are given and they have to be grouped with texts from the same author; we also do verification, which is similar to attribution but we opted to fine-tune the model with a siamese network to check if a texts has been written by the same person. Our reddit test set contains several authors whose texts have been shuffled and our objective is finding the author.

\subsection{Future works (\& the issue with topic)}
We first address the elephant in the room: What about topic? This is a common issue in style detection, as topic may bleed into the model contaminating results, that's why many challenges have started discriminating style from topic, usually restricting to narrow or even single topic. As can be read in this article, we do not control for topic which could be seen as a disadvantage or weakness of the model.

And indeed it is, but, our pretraining method is inherently author-oriented. Authors may be single-issue, write a lot about the same single topic among other problems, and no (author) dataset in existence and to the best of our knowledge, controls both: author and topic. Topic modelling could be performed on the inputs but that would add even more overhead to the already time-consuming training.

There has been a great deal of advances in disentangling style form content in computer vision, but not so much for text. Thus, this topic if solved would make a breakthrough in this kind of model, which is an obvious future line. 

Other possibilities for improvement are increasing the data heterogeneity with other datasets such as including fanfiction (Other kind of literary works), more exotic writing patterns (4chan-style forum postings) among many other styles of writing. The online aspect of this study poses another issue, do authors write the same way on twitter and reddit? It is obvious that they don't given each network presents different constraints for publication, thus it would be interesting how social network styles crossover and also how to control a dataset for this new factor.

\section*{Acknowledgments}
This work has been funded by the project PCI2022-134990-2 (MARTINI) of the CHISTERA IV Cofund 2021 program, funded by MCIN/AEI/10.13039/501100011033 and by the “European Union NextGenerationEU/PRTR”; by the research project DisTrack: Tracking disinformation in Online Social Networks through Deep Natural Language Processing, granted by Mobile World Capital Foundation; by the Spanish Ministry of Science and Innovation under FightDIS (PID2020-117263GB-I00); by MCIN/AEI/10.13039/501100011033/ and European Union NextGenerationEU/PRTR for XAI-Disinfodemics (PLEC2021-007681) grant, by Comunidad Autónoma de Madrid under S2018/TCS-4566 grant, by European Comission under IBERIFIER - Iberian Digital Media Research and Fact-Checking Hub (2020-EU-IA-0252); and by "Convenio Plurianual with the Universidad Politécnica de Madrid in the actuation line of Programa de Excelencia para el Profesorado Universitario".
%Bibliography
\bibliographystyle{unsrt}  
\bibliography{references}

\end{document}